# Homotopy Perturbation Method for Image Restoration and Denoising

Keyvan Yahya, Jafar Biazar, Hossein Azari, Pouyan Rafiei Fard

*Abstract*— The famous Perona-Malik (P-M) equation which was at first introduced for image restoration has been solved via various numerical methods. In this paper we will solve it for the first time via applying a new numerical method called Homotopy Perturbation Method (HMP) and the correspondent approximated solutions will be obtained for the P-M equation with regard to relevant error analysis. Through the implementation of our algorithm we will access some effective results which are deserved to be considered as worthy as the other solutions issued by the other methods.

## I. Introduction

SINCE the 90's, PDE based methods have been welcomed widely by vision researches in image restoration and the other applications like Image Segmentation, Motion Analysis and Image Classification and etc. These methods are divided into three categories namely the second order PDEs, the fourth order PDEs and the complex diffusion [1]-[4]. In 1990, Pietro Perona and Jitendra Malik in their common paper [5] for the first time introduced a parabolic partial differential equation which was called thereafter "Perona-Malik" and we abbreviated it by "PM". This equation could be posed rapidly as a new mathematical tool with the great capabilities to image restoring by convolving the original image with a Gaussian kernel. The original form of the equation in [5] have been shown as:

$$\frac{\partial u(x,y,t)}{\partial t} = \nabla \cdot \big(c(x,y,t)\nabla u(x,y,t)\big)$$
$$u(x,y,0) = u_0(x,y) \qquad (1)$$

where $c(x,y,t)$ is the diffusion factor which were suggested in the two forms of $c(x,y,t) = \frac{u}{(1+\frac{|\nabla u|^2}{k^2})}$

Keyvan Yahya is now an independent researcher in the field of computational neuroscience and applied mathematics (corresponding author to provide e-mail: keyvan.yahya@aol.com).
Jafar Biazar, is now with the Mathematics Department, University of Guilan, P.O.Box 1914, P.C. 4193833697, Rasht, Iran (e-mail: biazar@guilan.ac.ir.).
Hossein Azari is with Faculty of Mathematical Science, University of Shahid Beheshti P.O.Box 19835-389, Tehran, Iran (e-mail: h_azari@sbu.ac.ir).
Pouyan Rafiei Fard is now an independent researcher in the field of Neuromusicology and computational neuroscience. (email: rafieifard@ce.sharif.edu)

and $c(x,y,t) = \exp(-\frac{|\nabla u|^2}{(k^2)})$ and also $u_0$ is the original image should be restored.

Some people tried to solve the P-M by various numerical methods such as finite difference method [6], explicit finite volume method [7], total variation method [8] and etc. Each of these methods have been used regarding some different criterions in empirical implementations for example in edge conservation, speed of algorithm and how optimized the algorithms is. They also showed its considerable efficiency via comparison between former filter methods and these PDEs.

Through the studying of a non-linear parabolic equation of the following general form, we search for a new approximated solution:

$$\begin{cases} \frac{\partial u(t,x)}{\partial t} = g(|G_\sigma * \nabla u|)|\nabla u| div \frac{\nabla u}{|\nabla u|}, \\ \frac{\partial u(t,x)}{\partial N} = 0 \\ \frac{\partial u(0,x)}{\partial t} = u_0(x) \end{cases} \qquad (2)$$

where $u(t,x)$ is the solution of this PDE (restored image) we are searching for, $N$ is an outward Normal to domain $\Omega$ and $u(0,x)$ is the original image. The gray-level function $u(t,x)$ is depended on two parameters: the scale parameter denoted by $t$ and the spatial coordinate $x$. $G$ is smoothing Kernel (for instance, an additive Gaussian filter) and $g$ is a non-increasing Lipchitz function.

But to get more conceivable results, many authors manipulated this general form of restoration equation and expanded the P-M to propose different forms of this equation. For example, Lions and Alvarez offered an interesting non-linear form of restoration equation [6] or in [9], the Authors also gave a different parabolic equation which is considerable too or in [10]-[11], the work was focused on a complex form of diffusion equation and so on.

To reach our aim, we appoint to use the non-linear and modified form of P-M equation proposed by [12] because of its facility and vigor. The equation they proposed can be expressed as the following form:

$$\begin{cases} \frac{\partial u(t,x)}{\partial t} = -F(x, u, \nabla u, \nabla^2 u), \\ \frac{\partial u(t,x)}{\partial N} = 0, \quad in \ [0,t] \times \partial \Omega \\ \frac{\partial u(0,x)}{\partial t} = u_0(x), \quad in \ \Omega \end{cases} \quad (3)$$

where $F(x, u, \nabla u, \nabla^2 u) = -div\ (g(|G_\sigma * \nabla u|)|\nabla u|)$, $\Omega \subset \mathbb{R}^2$ is a rectangular domain, g is a non-increasing Lipchitz function satisfies this condition: $g(0) = 1$, $0 < g(s) \rightarrow 0$ for $s \rightarrow \infty$ and $G_\sigma \in C^\infty(\mathbb{R}^d)$ is a smoothing Kernel with compact support with:

$$\int_{\mathbb{R}^d} G_\sigma(x) = 1 \ \text{and} \ G_\sigma(x) \rightarrow \delta_z \ \text{as} \ \sigma \rightarrow 0. \quad (4)$$

$\delta_z$ is the Dirac delta function and because of boundary condition, we can assume that $u_0(x) \in W^{1,2}(\Omega)$ [13].

We are seeking for a solution belongs to a functional space which is the intersection of the Sobolov and the Lebesgue spaces ($L^1(\Omega)$). thus, we can consider $u$ as:

$$u \in W^{1,2}(\Omega) \cap L^1(\Omega), \quad (5)$$

As we mentioned before, most of the methods have been applied in order to find a solution follow a point by point process of approximation. In fact, Homotopy method has a more practical essence. To work with it, first of all we must construct *iterative series* in which we impose our PDE.

To start our study about the new method, let us express the explicit form of Perona-Malik equation used to restore the original image. The non-linear form of P-M equation can be expressed as the following:

$$\begin{cases} \frac{\partial u}{\partial t} = g(\|G_\sigma * \nabla u\|)|\nabla u| div\ \frac{\nabla u}{|\nabla u|}, \\ \frac{\partial u(t,x)}{\partial \eta} = o, \\ \frac{\partial u(0,x)}{\partial t} = u_0(x), \end{cases} \quad (6)$$

where $\eta$ is a normal outward to the domain $\Omega$, and as we know the function $u(t,x)$ is the solution of the nonlinear parabolic equation (restored image) we are seeking for based on $u(0,x)$ (original image). to work with this PDE, we are involved with three specific parameters, the first is the scale parameter denoted by $t$ (along with $x$ in spatial domain) and the second is called "Gaussian Kernel" sometimes we take it as the *additive Gaussian filter* and the latter $g(.)$ is an non-increasing Lipchitz function. It is the non linear one which enhances edges temporarily, before slowly blurring them out. It may be ill-posed, but can be made well-posed by replacing |r u| by |r uσ| where uσ is the convolution of $u$ by a Gaussian of standard deviation σ. However it is numerically well-posed, because of the implicit regularizing effect of numerical derivatives.

For numerical implementation we must adapt trivially the scheme proposed for the edge preserving linear diffusion instead of re-computing the diffusivity at each step. Homotopy Method is a Free Mesh numerical instrument which can give the solution. We put our PDE in the algorthim that we will offer below and gain a solution which is comparable with the others obtained via other mesh free methods. Also, the existence and uniqueness of the solution gained by homotopy method was proved earlier [14]. Of course, many authors see the homotopy method as a tool for finding an appropriate starting point $x^0$ in Newtonian method which is still a crucial problem. To use the homotopy method, it is necessary to construct a *parameter depending function* along side with additional conditions

## II. BASIC IDEA OF HOMOTOPY PERTURBATION METHOD

Homotopy perturbation method (HPM) has been proposed by "He" in 1998 [14], and later on, He himself developed and improved this method and applied it to boundary value problem [17], non-linear wave equations [19], and many other problems [16]–[19]. Homotopy perturbation method can be considered as the one which is universally capable to solve various kinds of non-linear functional equations. For example, it was also applied to non-linear Schrodinger equations [20], to a system of Volterra integral equations of the second kind [21], to the generalized Hirota–Satsuma coupled KDV equation [24], and to other equations [15]-[25].

In this method, the solution is taken into account as the summation of an infinite series which usually converge rapidly to the exact solution. To illustrate the basic concept of HPM, consider the following nonlinear differential equation:

$$A(u) - f(r) = 0, \quad r \epsilon \ \Omega \quad (7)$$

with boundary conditions :

$$B\left(u, \frac{\partial u}{\partial n}\right) = 0, \quad r \epsilon \ \Gamma \quad (8)$$

where $A$ is a general differential operator, $B$ is a boundary operator, $f(r)$ is a known analytic function, and $G$ is the boundary of the domain $W$. Generally speaking, the operator $A$ can be divided into two parts $L$ and $N$, where $L$ is a linear operator while $N$ is a nonlinear one. therefore, (7) can be rewritten as following:

$$L(u) + N(u) - f(r) = 0, \quad (9)$$

We construct a Homotopy $v(r,p): \Omega \times [0,1] \rightarrow \mathbb{R}$ which satisfies:

$$H(v,p) = (1-p)[L(v) - L(u_0)] + p[A(v) - f(r)] = 0, \quad (10)$$

Or:

$$H(v,p) = L(v) - L(u_0) + pL(u_0) + p[N(v) - f(r)] = 0 \quad (11)$$

where $p \in [0,1]$ is an embedding parameter, $u_0$ is an initial approximation for the solution of (7), which satisfies the boundary conditions. According to the HPM, we can first use the embedding parameter $p$ as a small parameter, and assume that the solution of (11) can be written as a power series in $p$:

$$v = v_0 + v_1 p + v_2 p^2 + \cdots, \quad (12)$$

setting $p = 1$, results in the approximate solution of (7)

$$u = \lim_{p \to 1} v = v_0 + v_1 + v_2 + \cdots \qquad (13)$$

III. SOLUTION OF THE PERONA-MALIK EQUATION BY HPM

Assume the following equation with indicated initial condition:

$$\begin{cases} u_t = \frac{1}{u_x^2 + u_y^2}\left(u_y^2 u_{xx} - 2u_x u_y u_{xy} + u_x^2 u_{yy}\right) \\ u(x,y,0) = \sqrt{x^2 + y^2} - 1 \end{cases} \qquad (14)$$

According to the HPM, the following homotopy can be constructed:

$$H(v,p) = (1-p)\left[\frac{\partial v}{\partial t} - \frac{\partial u_0}{\partial t}\right]$$
$$+ p\left[\frac{\partial v}{\partial t}\right.$$
$$+ \frac{1}{v_x^2 + v_y^2}\left(v_y^2 v_{xx} - 2v_x v_y v_{xy} + v_x^2 v_{yy}\right)$$
$$\left. - \frac{\partial u_0}{\partial t}\right] = 0 \qquad (15)$$

$$\frac{\partial v}{\partial t} - \frac{\partial u_0}{\partial t} = p\left(-\frac{1}{v_x^2 + v_y^2}\left(v_y^2 v_{xx} - 2v_x v_y v_{xy} + v_x^2 v_{yy}\right) - \frac{\partial u_0}{\partial t}\right). \qquad (16)$$

To solve (16) by HPM, Let us take the solution $v$ as the summation of the series (12). We suppose that:

$$v_0(x,y,t) = u(x,y,0) = \sqrt{x^2 + y^2} - 1 \qquad (17)$$

Substituting (12) into (16), using Maclaurin expansion and comparing the coefficients of identical degrees of $P$, lead us to:

$$v_1(x,y,t) = \frac{t}{\sqrt{x^2 + y^2}}, \qquad (18)$$

$$v_2(x,y,t) = -\frac{1}{2}\frac{t^2}{\left(x^2 + y^2\right)^{\frac{3}{2}}},$$

$$v_3(x,y,t) = \frac{1}{2}\frac{t^3}{\left(x^2 + y^2\right)^{\frac{5}{2}}},$$

$$v_4(x,y,t) = -\frac{5}{8}\frac{t^4}{\left(x^2 + y^2\right)^{\frac{7}{2}}},$$
$$\vdots$$

A ten-term approximation of the solution can be considered as following:

$$u(x,y,t) \approx \sum_{i=0}^{9} v_i(x,y,t) \qquad (19)$$

$$= \sqrt{x^2 + y^2} - 1 + \frac{t}{\sqrt{x^2 + y^2}} - \frac{1}{2}\frac{t^2}{\left(x^2 + y^2\right)^{\frac{3}{2}}} + \frac{1}{2}\frac{t^3}{\left(x^2 + y^2\right)^{\frac{5}{2}}}$$

$$- \frac{5}{8}\frac{t^4}{\left(x^2 + y^2\right)^{\frac{7}{2}}} + \frac{7}{8}\frac{t^5}{\left(x^2 + y^2\right)^{\frac{9}{2}}} - \frac{21}{16}\frac{t^6}{\left(x^2 + y^2\right)^{\frac{11}{2}}}$$

$$+ \frac{33}{16}\frac{t^7}{\left(x^2 + y^2\right)^{\frac{13}{2}}} - \frac{429}{128}\frac{t^8}{\left(x^2 + y^2\right)^{\frac{15}{2}}} + \frac{715}{128}\frac{t^9}{\left(x^2 + y^2\right)^{\frac{17}{2}}}$$

The plots of the approximated solutions for $t = 0, 1, 10, 50$ are presented in Fig. 2 to Fig. 6. (See the Appendix). Finally, After implementation of the obtained solution based on Homotopy method, we get the results which are shown in the Fig.1.

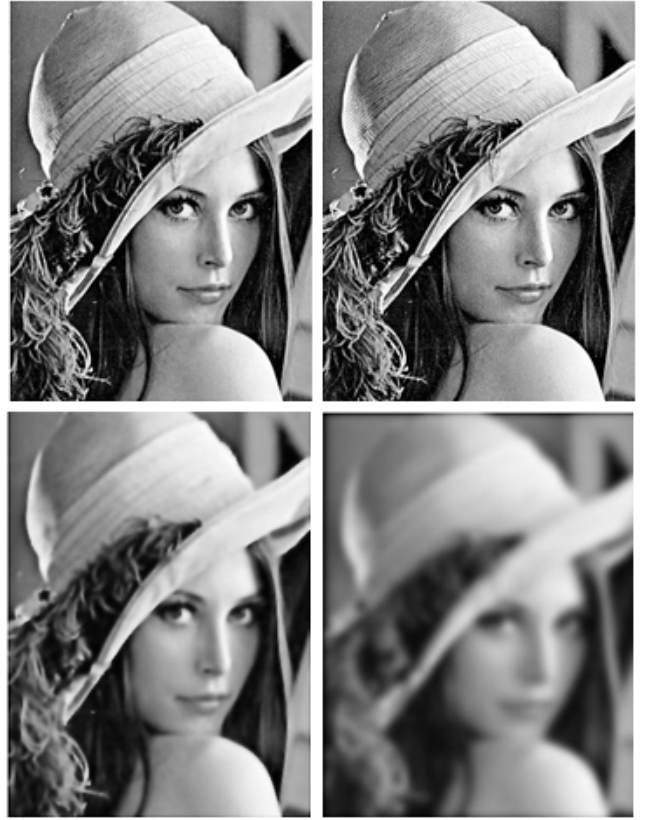

Fig.1. The approximated solutions of the P-M which were obtained via Homotopy method in image restoration.(top, from left to right) a) Lena's photo at the original image $t = 0$, b) restored photo after $t = 1$, c) after $t = 10$, d) after $t = 50$.

## IV. CONCLUSION

We could believe that there might be other methods which can solve the problem regularly. But we intend to examine a new numerical paradigm called "Level Set Method" in this problem. We expect that since the level set has some valuable properties such as reducing of dimensions, removing singularities and etc. by applying it we can obtain new optimized algorithm which will lead us to more speedy computations. We propose that by elaborating this new combined method, we can assign another application to the level set method that is image restoration. It might be used instead of usual methods which people have already applied.

## APPENDIX

Plots of the approximated solutions for $t = 0, 1, 10, 50$ are presented in Fig. 2 to Fig. 5.

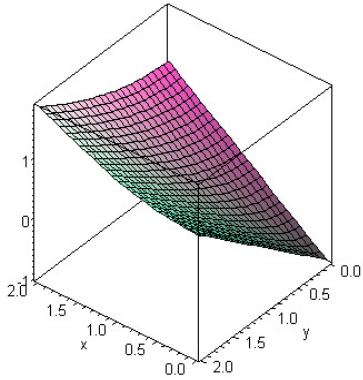

Fig. 2 HMP solution for t=0

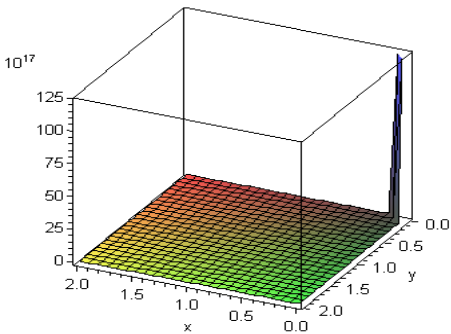

Fig. 3 HMP solution for t=1

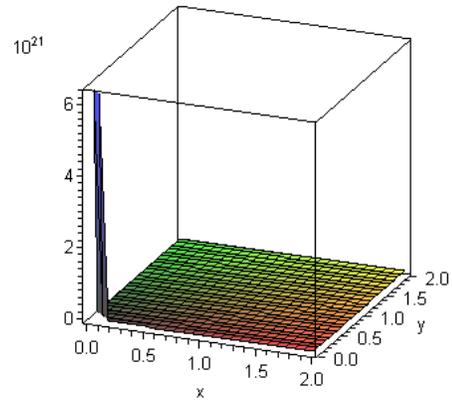

Fig. 4 HMP solution for t=10

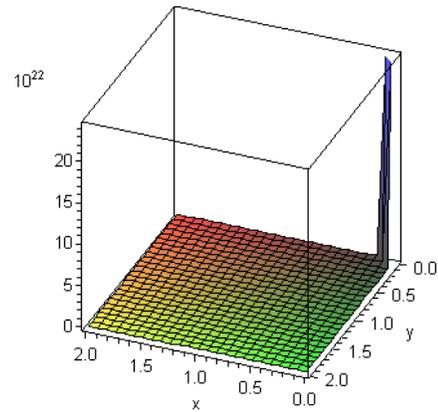

Fig. 5 HMP solution for t=50

## ACKNOWLEDGMENT

The Authors may wish to thank Zainab Ayati and Saeed Mehraban for their cooperation and also Mohsen Ghane-Basiri, Mahvash Almassian, Farahnaz Hassanzadeh and Fereshteh Zaker for their kind supports.